\documentclass[runningheads]{llncs}

\usepackage{eccv}

\usepackage{eccvabbrv}

\usepackage{graphicx}
\usepackage{booktabs}
\usepackage{multirow}

\usepackage[accsupp]{axessibility}  

\usepackage{hyperref}

\usepackage{orcidlink}

\usepackage{marvosym}

\begin{document}

\title{\texorpdfstring{D$^{2}$R$^{2}$OSR}{D2R2OSR}: 
Degradation-Disentangled Representation for Real-World \\ Omnidirectional Image Super-Resolution} 

\title{\texorpdfstring{D$^{2}$R$^{2}$OSR: Degradation-Disentangled Representation for Real-World \\ Omnidirectional Image Super-Resolution}{
D2R2OSR: Degradation-Disentangled Representation for Real-World Omnidirectional Image Super-Resolution
}
}

\titlerunning{D$^\text{2}$R$^\text{2}$OSR}

\author{Hongyu An\inst{1} \and
Xinfeng Zhang\inst{1}\thanks{Corresponding author.}\textsuperscript{(\Letter)} \and
Xu Fan\inst{1} \and
Shijie Zhao\inst{2} \and \\
Li Zhang\inst{2}\and
Ruiqin Xiong\inst{3}
}

\authorrunning{H. An et al.}

\institute{School of Computer Science and Technology, University of Chinese Academy of Sciences, Beijing, China \\
\email{\{anhongyu22, fanxu24\}@mails.ucas.ac.cn, xfzhang@ucas.ac.cn}
\and
ByteDance Inc., Shenzhen, China  \\ 
\email{\{zhaoshijie.0526, lizhang.idm\}@bytedance.com}
\and
School of Computer Science, Peking University, Beijing, China \\
\email{rqxiong@pku.edu.cn}
}

\maketitle

\vspace{-16pt}

\begin{abstract}
  With the growing demand for immersive visual experiences, high-quality omnidirectional images (ODIs) have become increasingly important.
  However, limitations in imaging devices and transmission bandwidth often lead to low-resolution ODIs, hindering the rendering of fine-grained 360$^{\circ}$ details, especially in the presence of real-world degradations and geometric distortions. 
  Existing real-world super-resolution (Real-SR) methods are inadequate for ODIs, as their degradation models fail to account for the complex imaging pipeline involving fisheye capture and Equirectangular Projection (ERP), introducing severe aliasing and projection-specific distortions.
  To address these challenges, we propose \textbf{D$^\text{2}$R$^\text{2}$OSR}, a \textbf{D}egradation-\textbf{D}isentangled \textbf{R}epresentation framework for \textbf{R}eal-world \textbf{O}mnidirectional image \textbf{S}uper-\textbf{R}esolution.
  D$^\text{2}$R$^\text{2}$OSR explicitly models degradations arising from both fisheye imaging and ERP projection, guided by two key insights: (1) projection priors play a critical role in shaping real-world degradations, and (2) human perception in immersive environments is inherently viewpoint-centric.
  Accordingly, we introduce a Perspective Projection Representation (PPR) operating alongside the ERP branch to capture viewpoint-aware features, together with a Degradation-Specific Module (DSM) that jointly models ERP-induced geometric distortions and PPR-specific real-world degradations.
  Extensive experiments demonstrate that D$^\text{2}$R$^\text{2}$OSR achieves state-of-the-art performance and produces visually compelling, high-fidelity omnidirectional Real-SR results while maintaining favorable computational efficiency for low-resource deployment.
\end{abstract}

\section{Introduction}
\label{sec:intro}
Omnidirectional images (ODIs), also known as 360$^{\circ}$ or panoramic images, capture the entire 360$^{\circ}$ field of view (FoV), enabling realistic visual experiences. 
With the rapid development of virtual and augmented reality applications, the demand for high-quality ODIs has increased dramatically.
To ensure immersion, ODIs must be rendered at ultra-high resolutions (e.g., 4K, 8K, or even 16K), while local regions of visual attention—typically spanning 100$^{\circ}\times100^{\circ}$ to 120$^{\circ}\times120^{\circ}$ FoV \cite{100viewpoint}—should remain sharp and detailed.
However, most ODIs are constrained by limited resolution owing to the high cost of precision imaging systems and restricted transmission bandwidth.
Moreover, throughout the practical imaging pipeline—including acquisition, stitching, projection, transmission, processing, and display—ODIs suffer from diverse degradations such as blur, noise, resize, compression, and projection-induced distortions, all of which severely impair visual fidelity.
Addressing these degradations through faithful high-quality reconstruction remains a fundamental yet unresolved challenge.

Super-resolution (SR) has therefore emerged as an effective client-side solution for enhancing ODI quality.
Nevertheless, conventional SR methods \cite{SRCNN,EDSR,SRGAN,RCAN,HAT,IPG} are typically trained under the idealized Bicubic down-sampling assumption, which deviates significantly from real-world degradation processes and limits their generalization.
Recent real-world SR (Real-SR) approaches \cite{RealSR, DASR_, BSRGAN, Real-ESRGAN, DASR__, DASR, RealBasicVSR} attempt to mitigate this gap by either implicitly learning degradation priors from low-resolution data or explicitly synthesizing realistic degradation models.
Despite their progress, these methods largely overlook the mixed degradations and projection-related distortions intrinsic to the ODI imaging pipeline, where multiple degradation sources jointly hinder the recovery of photo-realistic, high-fidelity omnidirectional images.

\begin{figure*}[!t]
\centering
\includegraphics[width=\columnwidth]{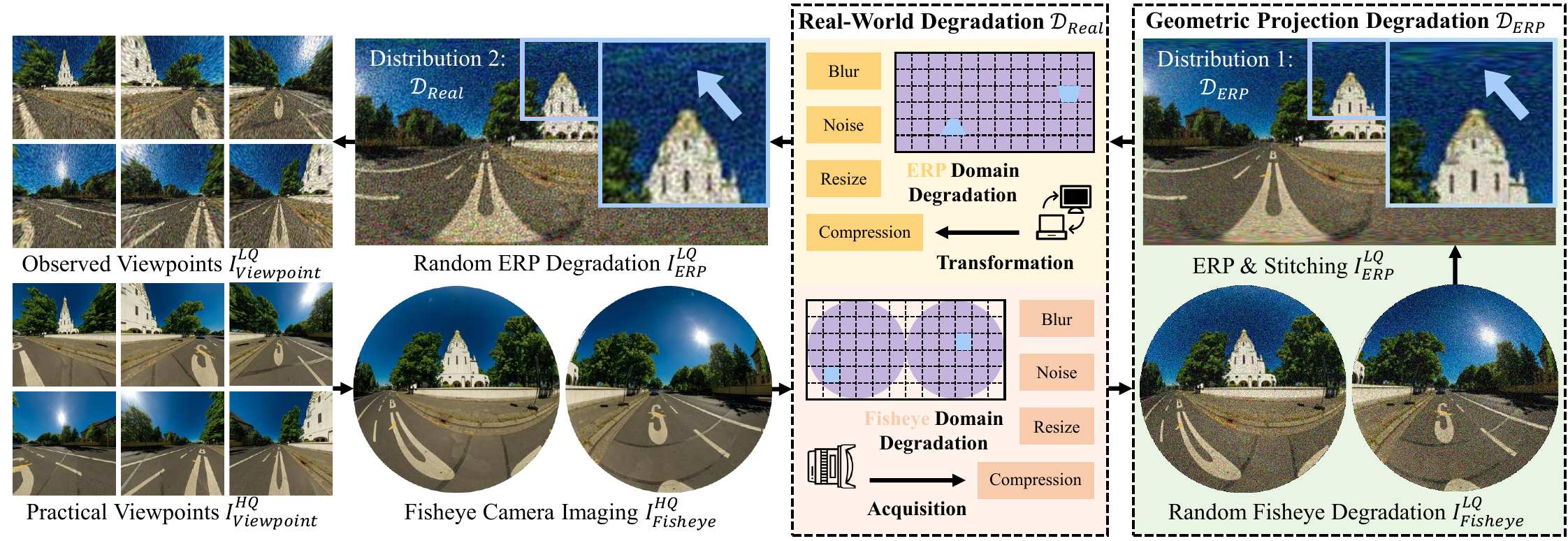}
\caption{Real-world degradations in the omnidirectional imaging pipeline.
During fisheye capture and subsequent ERP conversion, degradations such as blur, noise, resize, and compression are introduced and further amplified by projection-induced distortions (e.g., geometric stretching and shape deformation), resulting in severe visual artifacts.
The zoomed-in regions reveal distinct, domain-specific degradation characteristics in fisheye and ERP domains, motivating separate modeling for faithful restoration.}
\label{fig:degradation}
\end{figure*}

ODIs are typically captured using fisheye cameras and stitched to cover the full 360$^{\circ}$ FoV.
To ensure compatibility with general image transmission and processing pipelines, the spherical content is projected onto planar representations, where pixel-wise mapping inevitably introduces information loss and interpolation distortions.
Among various projection methods, Equirectangular Projection (ERP) is the most widely adopted representation for its computational efficiency and broad applicability, enabling direct processing as standard 2D images despite non-uniform content density.
After projection, ERP ODIs are subjected to compression, transmission, and enhancement, and are finally rendered as localized perspective views for immersive visualization.
As illustrated in Fig. \ref{fig:degradation}, diverse real-world degradations $\mathcal{D}_{Real}$ arise throughout the entire ODI imaging pipeline, while ERP further introduces intrinsic geometric distortions $\mathcal{D}_{ERP}$, together yielding complex and mixed degradation distributions.
Existing ODI-SR methods \cite{360-SS, LAU-Net, LAU-Net+, ICIP, OSRT, FATO, OmniSSR, BPOSR, ODA-SRN, GDGT-OSR, FAOR} primarily focus on mitigating latitude-dependent distortions by leveraging geometric priors in ERP space.
Although these approaches partially alleviate projection-related degradations, they remain insufficient for addressing the full spectrum of real-world degradations encountered in practical ODI imaging pipelines, ultimately limiting restoration quality and generalization capability.

To address the challenges of real-world ODI-SR, we propose a mixed fisheye–ERP degradation model that synthesizes realistic training samples by capturing diverse degradation distributions across the imaging pipeline.
Building on this, we develop \textbf{D$^\text{2}$R$^\text{2}$OSR} (\textbf{D}egradation-\textbf{D}isentangled \textbf{R}epresentation-Based \textbf{R}eal-World \textbf{O}mnidirectional Image \textbf{S}uper-\textbf{R}esolution). 
It is a unified framework that disentangles latitude-dependent projection distortions from blind pixel-level degradations.
Specifically, a perspective-aware representation branch is introduced alongside the ERP branch to enhance local viewpoint quality, while degradation-specific encoding enables both branches to adaptively compensate for distortions with or without explicit spatial priors.
A projection fusion mechanism further integrates complementary features across the two domains.
The main contributions of this paper are summarized as follows:
\begin{enumerate}
\item We present a unified degradation modeling paradigm for real-world ODI-SR that captures the coupled effects of fisheye imaging and ERP projection throughout the omnidirectional imaging pipeline.
By explicitly disentangling degradations with distinct priors, D$^\text{2}$R$^\text{2}$OSR enables more faithful and grounded restoration of real-world ODIs.
\item We propose a Perspective Projection Representation (PPR) that bridges spherical and planar domains, effectively decoupling projection-induced geometric distortions from pixel-level real-world degradations and facilitating accurate degradation modeling from a viewpoint-centric perspective.
\item We develop a dual-domain ODI-SR framework that jointly leverages complementary omnidirectional and perspective representations, equipped with Degradation-Specific Modules (DSMs) for adaptive degradation compensation to robustly handle diverse and mixed degradation scenarios.
\end{enumerate}

\section{Related Work}
\label{sec:rela}

\subsection{Real-World Image Super-Resolution}
With the rapid advances in deep learning, network-based image SR methods \cite{SRCNN,EDSR,SRGAN,RCAN,HAT,IPG} have largely replaced traditional approaches.
However, most existing models are trained under predefined degradation assumptions with fixed patterns, which fail to capture the complexity and diversity of real-world degradations.
In practice, low-resolution (LR) images are acquired under varying devices and environments, and are commonly affected by blur, noise, resize, and compression.
To bridge the gap between synthetic and real-world degradations, a series of works have explored real-world SR (Real-SR).
RealSR \cite{RealSR} constructs a large kernel pool to cover diverse degradation patterns.
BSRGAN \cite{BSRGAN} and Real-ESRGAN \cite{Real-ESRGAN} explicitly design mathematical degradation pipelines for realistic data synthesis, while ADL \cite{ADL} simulates unknown down-sampling processes via adversarial learning.
DASR \cite{DASR} adaptively estimates degradation from LR inputs to modulate network parameters, and LWay \cite{LWay} combines supervised pre-training with self-supervised learning to mitigate the synthetic–real domain gap.
DKP \cite{DKP} proposes unsupervised kernel estimation via Markov chain Monte Carlo sampling, whereas AdaSR \cite{AdaSR} leverages sample-adaptive priors learned from image self-similarity.
More recently, diffusion-based models \cite{DiffBIR, SeeSR, SUPIR} have further improved perceptual realism by introducing powerful generative priors.
Nevertheless, these methods incur substantial computational overhead due to their diffusion backbones, which limits their scalability to ultra-high-resolution scenarios such as omnidirectional images.

\subsection{Omnidirectional Image Super-Resolution}
In ODI applications, ERP remains the dominant representation due to its simplicity, full-sphere coverage, and strong compatibility with standard architectures.
Consequently, most existing ODI-SR methods \cite{360-SS, LAU-Net, ICIP, OSRT, GDGT-OSR, FATO, BPOSR, ODA-SRN, OmniSSR, STDAN} are developed in the ERP domain and focus on mitigating geometric distortions introduced by spherical-to-planar projection.
Early works \cite{360-SS, LAU-Net} mitigate latitude-dependent distortions and uneven content distribution through projection-related modulation.
\cite{ICIP} introduces a perception-oriented network to enhance high-frequency details from critical viewpoints.
Subsequent studies explore more advanced distortion modeling and representation strategies.
OSRT \cite{OSRT} aligns features using continuous and adaptive offsets to alleviate projection-induced misalignment, while GDGT-OSR \cite{GDGT-OSR} extends this idea with distortion-guided rectangle-window self-attention for improved texture modeling.
FATO \cite{FATO} addresses content density imbalance by leveraging frequency-domain characteristics and high-frequency attention.
BPOSR \cite{BPOSR} combines ERP and cubemap projections through parallel attention to exploit complementary representations, whereas ODA-SRN \cite{ODA-SRN} employs multi-segment parameterized convolutions for geometric distortion compensation.
OmniSSR \cite{OmniSSR} introduces octadecaplex tangent interaction and gradient decomposition for zero-shot ODI-SR.
Despite these advances, existing methods primarily assume fixed Bicubic down-sampling and emphasize ERP geometric distortions, overlooking the complex real-world degradation encountered in ODI imaging pipelines and the non-uniform human visual perception across local viewpoints.
These limitations impair practical applicability and immersive visual quality.

\begin{figure*}[!t]
\centering
\includegraphics[width=\columnwidth]{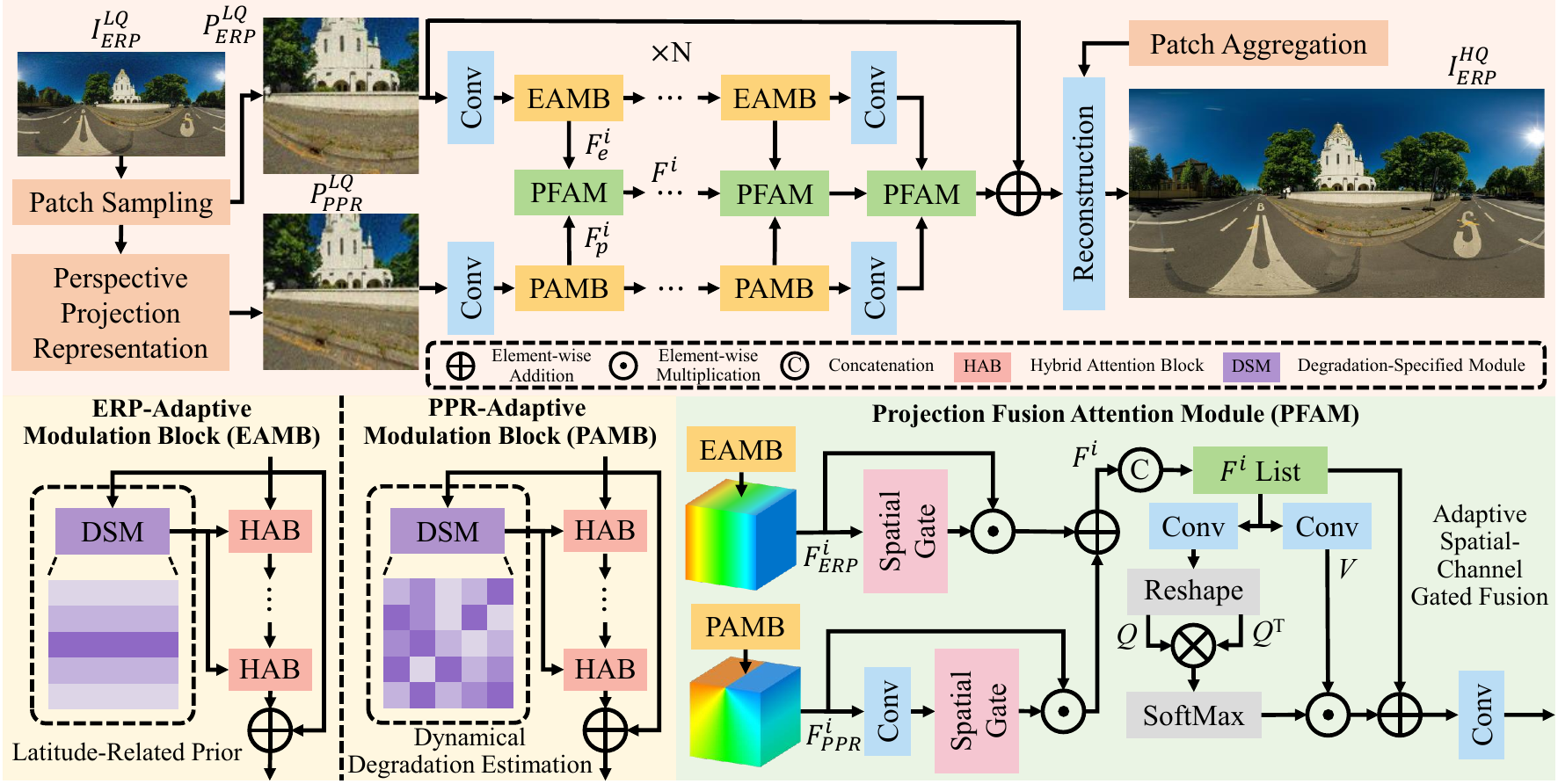}
\caption{Overall framework of D$^\text{2}$R$^\text{2}$OSR.
It comprises two parallel branches, ERP and PPR, for distortion-aware restoration.
PPR patches are derived from ERP patches in a nearly lossless manner, preserving consistent scene content.
The PPR branch captures viewpoint-centric features to enhance perceptual quality.
DSMs in the ERP branch compensate for projection-induced geometric distortions, while those in the PPR branch handle local real-world degradations.
PFAMs fuse complementary features across branches to facilitate information exchange.}
\label{fig:framework}
\vspace{-2pt}
\end{figure*}

\section{Method}
\label{sec:method}

\subsection{Overall Framework}

The D$^\text{2}$R$^\text{2}$OSR pipeline is demonstrated in Fig. \ref{fig:framework}.
Given a low-quality (LQ) ERP image $I_{ERP}^{LQ}$ with geometric and real-world degradations, our goal is to learn an adaptive model $\mathcal{M}$ that restores high-resolution details by modulating $\mathcal{D}_{ERP}$ and $\mathcal{D}_{Real}$.
Formally, $I_{ERP}^{LQ}$ is partitioned into ERP patches $P_{ERP}^{LQ}$ and transformed into perspective patches $P_{PPR}^{LQ}$ via Perspective Projection Representation (PPR), facilitating the disentanglement of ERP-induced and real-world degradations.
Both patch sets are encoded into shallow features $F_{e}^{0}$ and $F_{p}^{0}$ and processed by parallel ERP and PPR branches, each consisting of $N$ EAMBs and PAMBs with embedded DSMs.
DSMs in EAMBs model latitude-related $\mathcal{D}_{ERP}$, while those in PAMBs focus on $\mathcal{D}_{Real}$.
PFAMs fuse $F_{e}^{i}$ and $F_{p}^{i}$ at each stage, progressively refining complementary features for the final reconstruction.

\subsection{Perspective Projection Representation (PPR)}


\begin{figure}[!t]
\centering
\includegraphics[width=0.95\columnwidth]{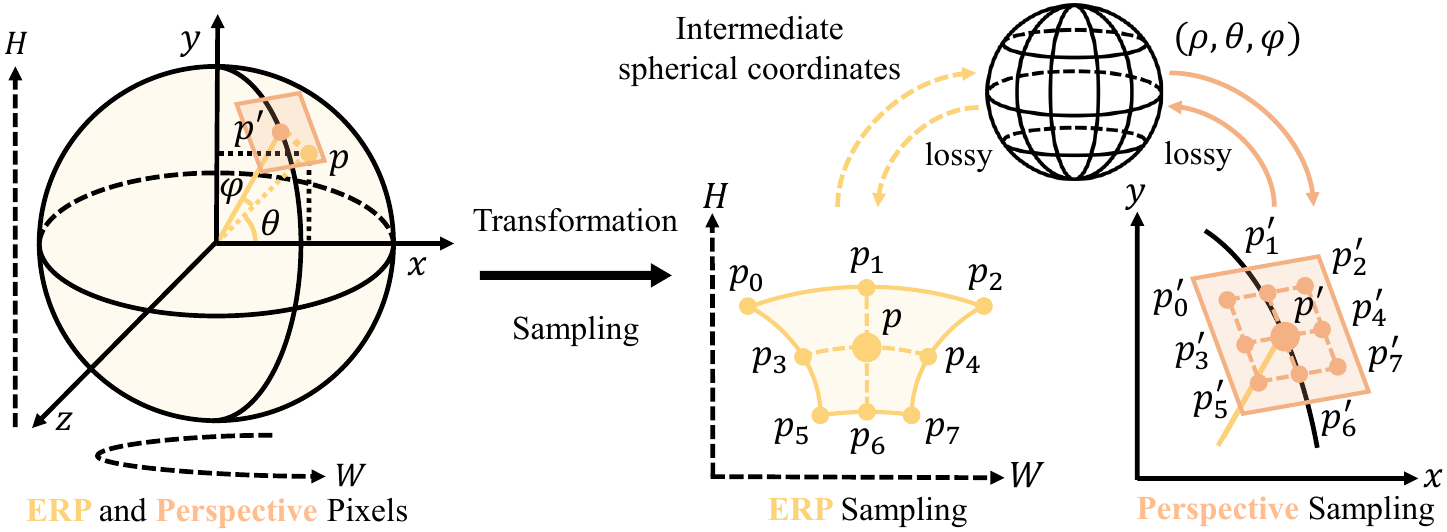}
\caption{Illustration of pixel distributions in the Equirectangular Projection (ERP) and Perspective Projection Representation (PPR) domains.
ERP exhibits increasing latitude-dependent pixel stretching, whereas PPR maintains a uniform pixel distribution and better preserves local geometry.}
\label{fig:PPR0}
\end{figure}

Although ERP is the dominant representation for ODIs due to its simplicity and broad compatibility, projecting spherical scenes onto a planar domain inevitably introduces severe geometric distortions, particularly at high latitudes.
When these distortions interact with real-world degradations throughout the ODI imaging pipeline (Fig.~\ref{fig:degradation}), faithful content restoration becomes substantially more challenging.
To address this issue, we introduce a perspective projection representation (PPR) that processes features in a projection-independent, viewpoint-centric domain better aligned with human visual perception.

As shown in Fig. \ref{fig:PPR0}, the ERP-PPR transformation is formulated along the spherical cutting-plane direction.
Specifically, ERP coordinates are first mapped onto the unit sphere, followed by a rotation about the sphere center $o$ to simulate camera motion.
The tangent plane at the rotated viewpoint (orange region) is then projected onto the image plane, forming an ERP$\to$Sphere$\to$PPR pipeline. 
Conventional interpolation-based resampling (e.g., Bicubic or Bilinear interpolation) relies on intermediate spherical warping, inevitably introducing information loss and blurring artifacts.
To this end, we construct a more accurate perspective representation that decouples projection-included degradations from real-world distortions.
Formally, for a target PPR coordinate $p'(u',v')$ corresponding to an ERP coordinate $p(u,v)$, we define a continuous mapping $p'=f(p)$.
The local offset $\delta p'$ is estimated from $p$ and its neighborhood:
\begin{equation}
\mathcal{P}=\{p_i|p_i=f^{-1}(p')+[\frac{m}{h},\frac{n}{w}],[m,n]\in[-1,0,1]\},
\end{equation}
where $h$ and $w$ denote the height and width of the ERP representation, respectively.
Under a first-order Taylor expansion, the mapping can be locally approximated in a linear form as follows:
\begin{equation}
\begin{aligned}
\delta p_i'&=f(p)-f(p_i)\approx f(p_i)+J_{f}(p_i)(p-p_i)-f(p_i) \\
&=J_{f}(p_i)(p-p_i)=J_{f}(p_i)\delta p,
\end{aligned}
\label{equ:1}
\end{equation}
where $J_{f}(p_i)$ denotes the Jacobian of $f$ at $p_i$, and $\delta p=p-p_i$ is the local grid in the ERP domain.
Since the ERP-to-PPR mapping varies with latitude, the first-order approximation cannot accurately characterize the spatially varying projection warping.
To better model these non-uniform distortions, we further introduce higher-order geometric information.
Following LTEW\cite{LTEW} and ResVR\cite{ResVR}, we incorporate the second-order Hessian $H_{f}(\mathcal{P})$ together with $J_{f}(\mathcal{P})$ to jointly encode local orientation and curvature:
\begin{equation}
\begin{aligned}
J_{f}(\mathcal{P})&=\begin{bmatrix}
p_4'-p_3'\\
p_1'-p_6'
\end{bmatrix}, \\
H_{f}(\mathcal{P})&=\begin{bmatrix}p_3'+p_4'-2p'&p_2'+p_5'-p_0'-p_7'\\p_2'+p_5'-p_0'-p_7'&p_1'+p_6'-2p'
\end{bmatrix}.
\end{aligned}
\end{equation}

\begin{figure}[!t]
\centering
\includegraphics[width=0.95\columnwidth]{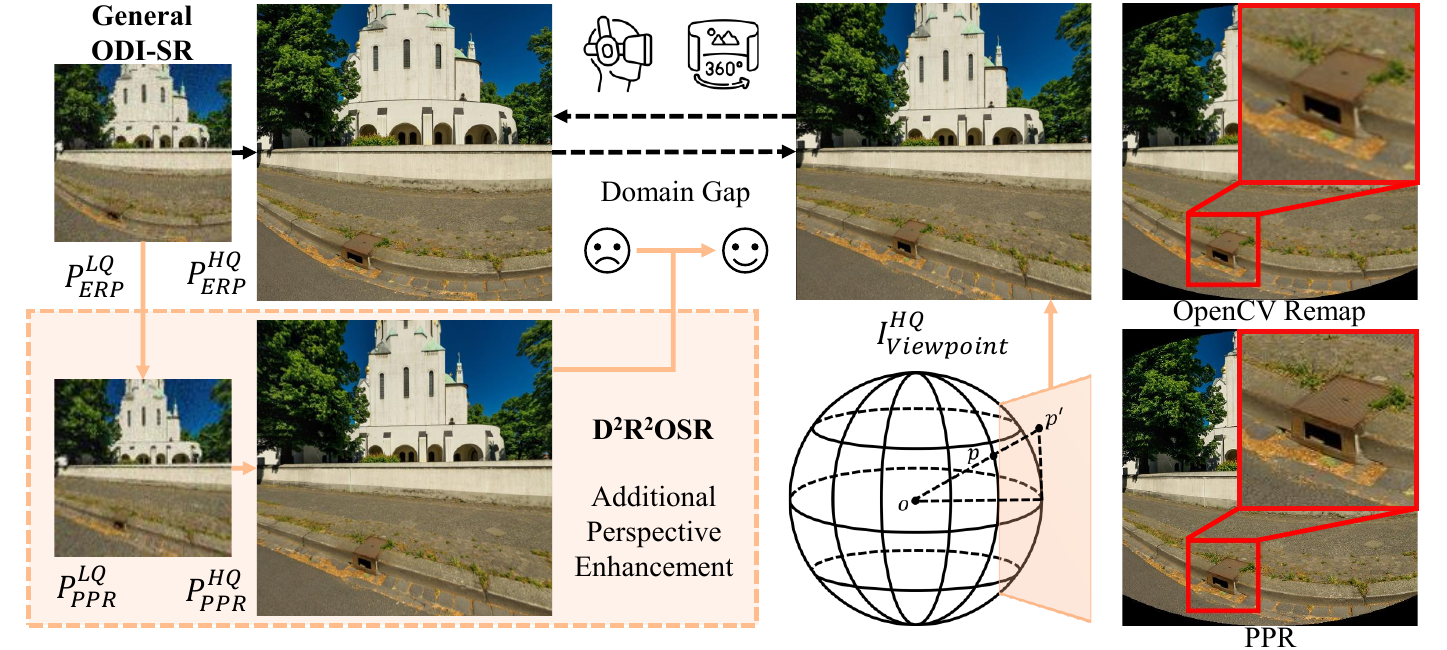}
\caption{PPR transforms ERP patches into perspective patches, aligning the viewing distribution with natural images and effectively decoupling projection-included degradations.
Compared with the conventional ERP-to-Perspective-to-ERP OpenCV remapping, PPR yields sharper and more structurally faithful reconstructions.}
\label{fig:PPR1}
\end{figure}

Moreover, the local displacement and its gradient in the PPR domain are approximated as $\delta p'=J_{f}(\mathcal{P})\delta p,$ and $\delta (\nabla p')=H_{f}(\mathcal{P})\delta(\nabla p)$.
Unlike prior works \cite{LTEW, ResVR}, which primarily utilize Jacobian and Hessian matrices for pixel-level warping, we explicitly formulate the ERP$\to$PPR conversion through $\delta p'$ and $\delta (\nabla p')$, establishing a continuous coordinate mapping framework.
Concretely, the six independent elements derived from $J_{f}$ and $H_{f}$ are concatenated to form spatial priors $s(\mathcal{P})$, enabling direct ERP$\to$PPR mapping without intermediate spherical resampling.
Inspired by implicit neural representation methods \cite{LIIF,SRWarp,LTEW}, we further design a learnable coordinate transformation estimator $E={E_a,E_f,E_p}$ conditioned on $s(\mathcal{P})$. 
This estimator models local textures in the Fourier domain, enabling adaptive reconstruction of high-frequency details.
The estimation is formulated as follows:
\begin{equation}
E(\mathcal{E}(p),\delta p,s(\mathcal{P}))=E_a(\mathcal{E}(p))\otimes\begin{bmatrix}\text{cos}\{\pi(<E_f(\mathcal{E}(p)),\delta p>+E_p(s(\mathcal{P})))\}\\\text{sin}\{\pi(<E_f(\mathcal{E}(p)),\delta p>+E_p(s(\mathcal{P})))\}
\end{bmatrix},
\label{equ:3}
\end{equation}
where $\mathcal{E}(\cdot)$ indicates the feature encoder, while $E_a$, $E_f$, and $E_p$ estimate the amplitude, frequency, and phase components, respectively.
Here, $<\cdot,\cdot>$ means the inner product.
Based on the learned coordinate transformation estimator, the PPR-domain pixel distribution is predicted as follows:
\begin{equation}
P^{LQ}_{PPR}=\sum_{i\in\mathcal{P}}\omega_i\mathcal{D}(E(\mathcal{E}(p),\delta p, s(p_i))),
\label{equ:4}
\end{equation}
where $\mathcal{D}(\cdot)$ denotes a four-layer MLP, and $\omega_i$ is a local ensemble coefficient.
As depicted in Fig. \ref{fig:PPR1}, compared with interpolation-based OpenCV Remap, the proposed formulation significantly reduces information loss during ERP$\to$PPR conversion.
Consequently, the PPR branch effectively decouples projection-induced geometric degradations, improves real-world degradation modeling, and better aligns with human visual perception.

\begin{figure}[!t]
\centering
\includegraphics[width=\columnwidth]{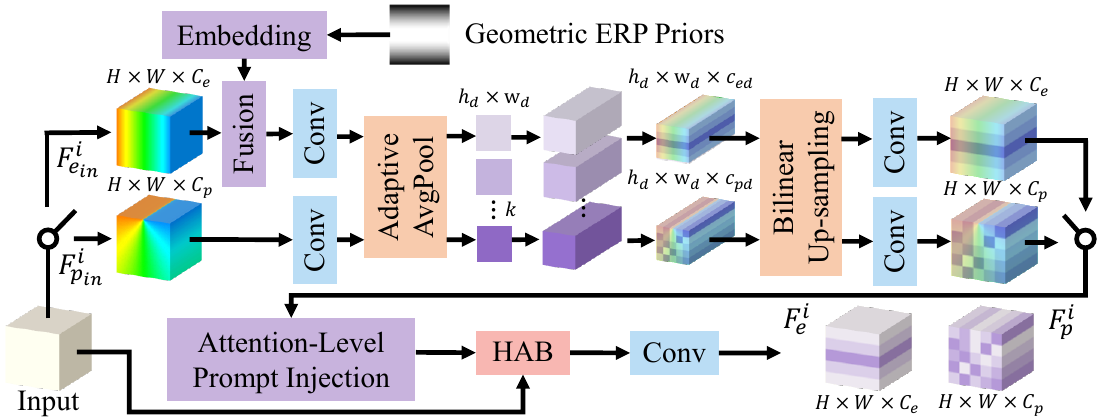}
\caption{Degradation-Specific Module (DSM). The DSM learns projection-aware adaptive guidance to compensate for projection-induced degradations.}
\label{fig:DSM}
\end{figure}

\subsection{Degradation-Specific Module (DSM)}

As illustrated in Fig. \ref{fig:framework}, we introduce two complementary modules: the ERP-Adaptive Modulation Block (EAMB) and the PPR-Adaptive Modulation Block (PAMB), to address multi-degradation ODI restoration challenges.
The EAMB operates in the ERP domain to mitigate latitude-dependent geometric distortions, while the PAMB models random real-world degradations in the perspective domain.
To exploit domain-specific priors, we embed Degradation-Specific Modules (DSMs) into both branches, enabling degradation-aware modulation under heterogeneous projection distributions.

Fig. \ref{fig:DSM} shows the DSM architecture. 
At the $i$th stage, the inputs to EAMB and PAMB are $F_{e}^{i} \in \mathbb{R}^{H\times W\times C_e}$ and $F_{p}^{i} \in \mathbb{R}^{H\times W\times C_p}$, respectively.
The DSM takes three inputs: branch features from EAMB/PAMB, a projection-aware conditional map, and a learnable degradation descriptor (``Prompt'').
Since feature distributions vary spatially across projections, we first extract projection-aware degradation priors using $1 \times 1$ convolutions.
For the ERP branch, a latitude-related quantization map is derived to characterize spatially varying geometric distortions, while a detailed derivation is provided in the Appendix.
The resulting conditional map of size $H\times W\times1$ is fused with $F_{e}^{i}$ to produce geometrically calibrated features $F_{e}^{i'}$.
To achieve distribution-adaptive modulation, we further design a lightweight encoder to generate learnable degradation descriptors $D_{ERP}^{i} \in \mathbb{R}^{h_d\times w_d\times c_{ed}}$ and  $D_{PPR}^{i} \in \mathbb{R}^{h_d\times w_d\times c_{pd}}$.
Specifically, features are processed by $3 \times3$ convolutions with LeakyReLU activations, followed by Adaptive Average Pooling (AAP) to obtain compact descriptor vectors.
These descriptors are subsequently up-sampled and refined via convolution layers to generate feature-level modulation masks. 
The generation of degradation-specific features in the ERP and PPR branches is summarized as:
\begin{equation}
DSF^i_{ERP/PPR}=\sum_{k=1}^{K} {\text{AAP}(\text{Conv}(F^{i'}_{e}/F^{i}_{p}))_kP_k},
\label{equ:5}
\end{equation}
where $k$ represents the descriptor length.
The resulting degradation-specific feature $DSF^i_{ERP/PPR}$ is concatenated with the branch input and subsequently fed into the HAB block~\cite{HAT} for degradation-aware feature modulation:
\begin{equation}
\begin{aligned}
F^{i'}=\text{Cat}(F^i_{e/p}, \text{Up}(\text{Conv}(DSF^i_{ERP/PPR}))), \ 
F^i_{e/p}=\text{Conv}(\text{HAB}(F^{i'})).
\end{aligned}
\end{equation}

The framework is optimized end-to-end using only the reconstruction objective, without explicit degradation annotations.
Built upon the representation-level disentanglement provided by PPR, the learnable descriptors implicitly organize different degradation characteristics into distinct feature subspaces.
To further enhance degradation-aware modulation, we inject prompts into the value branch of window-based multi-head self-attention as $V' = V + P_v$, where $P_v$ denotes a prompt tensor aligned with the multi-head structure.
By modulating only the value component, the proposed strategy preserves query-key similarity estimation while enabling degradation-aware contextual aggregation.
Compared with feature concatenation or bias-based conditioning, the proposed strategy enables head-wise contextual modulation without introducing additional attention branches or noticeable computational overhead.

\subsection{Projection Fusion Attention Module (PFAM)}
The aggregation of enhanced features across different distributions and stages is critical for ODI-SR performance.
As demonstrated in Fig. \ref{fig:framework}, EAMBs and PAMBs operate in distinct projection domains and hierarchical levels, making effective feature integration non-trivial.
Existing work \cite{BPOSR} directly concatenates branch features followed by attention-based fusion.
However, the distinct spatial distributions of ERP and PPR limit the effectiveness of naive concatenation.
To address this issue, we propose a Projection Fusion Attention Module (PFAM) for adaptive cross-branch fusion.

To effectively integrate ERP and PPR features from distinct projection domains, we propose an adaptive spatial-channel gated fusion mechanism, as illustrated in Fig.~\ref{fig:framework}.
Given ERP and PPR features $F_{e}^i \in \mathbb{R}^{B \times C_e \times H \times W}$ and $F_{p}^i \in \mathbb{R}^{B \times C_p \times H \times W}$, we first align channel dimensions via a learnable $1 \times 1$ convolution $\mathcal{P}: \mathbb{R}^{C_p} \rightarrow \mathbb{R}^{C_e}$, yielding $\tilde{F}_{p}^i = \mathcal{P}(F_{p}^i)$.

A branch-aware spatial attention map is then computed as:
\begin{equation}
S_{e/p} = \text{Sigmoid} \left( W_2 * \text{LeakyReLU} \left( W_1 * F_{e}^i / \tilde{F}_{p}^i \right) \right),
\end{equation}
where $W_i$ denotes convolution layers, and the resulting spatial gates assign adaptive importance to different projections.
The fusion is performed residually:
\begin{equation}
F^i = S_e \odot F_{e}^i + S_p \odot \tilde{F}_{e}^i.
\end{equation}
The fused feature set $\mathcal{F}=\{F^i\}$ spans multiple network stages.
To capture inter-stage dependencies, we further introduce a depth-wise attention mechanism over $\mathcal{F}$.
Features are reshaped as $Q,V\in\mathbb{R}^{B\times D\times(CHW)}$ and depth attention is computed as $\text{Softmax}(QQ^\top)$.
This design enables adaptive fusion across stages, producing more structurally coherent representations.
Overall, PFAM bridges ERP and PPR branches by reducing feature misalignment and adaptively balancing geometric and real-world cues, leading to more robust real-world ODI-SR.

\begin{table}[!t]
    \caption{
    Quantitative real-world ODI-SR comparisons on Flickr360, ODI-SR, and SUN360 datasets under $\times4$, $\times8$, and $\times16$ settings. 
    Best in \textcolor{red}{\textbf{red}}, second-best in \textcolor{blue}{\underline{blue}}.
    }
    \label{tab:SR}

    \centering
    \footnotesize
    \renewcommand{\arraystretch}{0.95}
    \setlength{\tabcolsep}{7pt}

    \begin{tabular}{c|cc|cc|cc}
        \toprule

        \multirow{2}{*}{Method} &
        \multicolumn{2}{c|}{$\times4$} &
        \multicolumn{2}{c|}{$\times8$} &
        \multicolumn{2}{c}{$\times16$} \\

        \cmidrule{2-7}

        &
        \multicolumn{6}{c}{WS-PSNR$\uparrow$ / WS-SSIM$\uparrow$} \\

        \midrule

        \multicolumn{7}{c}{\textbf{Flickr360 Dataset}} \\
        \midrule

        360-SS\cite{360-SS}      &24.20&0.6198&16.98&0.5728&15.91&0.5844 \\
        Real-ESRGAN\cite{Real-ESRGAN} &24.30&0.6171&21.85&0.6037&19.84&0.5934 \\
        SwinIR\cite{SwinIR}      &25.16&0.6827&22.83&0.6347&21.19&0.6202 \\
        HAT\cite{HAT}         &25.19&0.6838&22.93&0.6354&21.36&0.6206 \\
        PromptIR\cite{PromptIR}    &25.20&0.6852&22.87&\textcolor{blue}{\underline{0.6376}}&21.40&\textcolor{blue}{\underline{0.6240}} \\
        OSRT\cite{OSRT}        &25.41&0.6886&22.93&0.6354&21.41&0.6211 \\
        BPOSR\cite{BPOSR}       &25.34&0.6864&22.88&0.6348&21.27&0.6212 \\
        AdaIR\cite{AdaIR}       &\textcolor{blue}{\underline{25.49}}&\textcolor{blue}{\underline{0.6912}}
        &\textcolor{blue}{\underline{22.96}}&0.6377
        &\textcolor{blue}{\underline{21.45}}&0.6238 \\
        D$^\text{2}$R$^\text{2}$OSR (Ours)
        &\textcolor{red}{\textbf{25.64}}&\textcolor{red}{\textbf{0.6934}}
        &\textcolor{red}{\textbf{23.24}}&\textcolor{red}{\textbf{0.6433}}
        &\textcolor{red}{\textbf{21.64}}&\textcolor{red}{\textbf{0.6249}} \\

        \midrule

        \multicolumn{7}{c}{\textbf{ODI-SR Dataset}} \\
        \midrule

        360-SS\cite{360-SS}      &23.30&0.6031&17.11&0.5562&15.97&0.5667 \\
        Real-ESRGAN\cite{Real-ESRGAN} &23.24&0.6001&21.52&0.5904&19.61&0.5785 \\
        SwinIR\cite{SwinIR}      &23.97&0.6537&22.32&0.6129&20.77&0.5996 \\
        HAT\cite{HAT}         &23.97&0.6541&22.41&0.6139&20.92&0.6000 \\
        PromptIR\cite{PromptIR}    &24.00&0.6555&22.37&\textcolor{blue}{\underline{0.6165}}&20.96&\textcolor{blue}{\underline{0.6039}} \\
        OSRT\cite{OSRT}        &24.22&0.6569&22.39&0.6135&20.96&0.6004 \\
        BPOSR\cite{BPOSR}       &24.14&0.6541&22.36&0.6132&20.82&0.6006 \\
        AdaIR\cite{AdaIR}       &\textcolor{blue}{\underline{24.26}}&\textcolor{blue}{\underline{0.6591}}
        &\textcolor{blue}{\underline{22.42}}&0.6158
        &\textcolor{blue}{\underline{21.00}}&0.6036 \\
        D$^\text{2}$R$^\text{2}$OSR (Ours)
        &\textcolor{red}{\textbf{24.37}}&\textcolor{red}{\textbf{0.6626}}
        &\textcolor{red}{\textbf{22.62}}&\textcolor{red}{\textbf{0.6210}}
        &\textcolor{red}{\textbf{21.18}}&\textcolor{red}{\textbf{0.6053}} \\

        \midrule

        \multicolumn{7}{c}{\textbf{SUN360 Dataset}} \\
        \midrule

        360-SS\cite{360-SS}      &23.02&0.5943&16.64&0.5728&15.14&0.5875 \\
        Real-ESRGAN\cite{Real-ESRGAN} &23.07&0.5922&21.23&0.5926&19.34&0.5793 \\
        SwinIR\cite{SwinIR}      &23.66&0.6439&22.04&0.6296&20.48&0.6176 \\
        HAT\cite{HAT}         &23.67&0.6444&22.08&0.6300&20.55&0.6177 \\
        PromptIR\cite{PromptIR}    &23.69&0.6464&22.07&0.6311&20.55&\textcolor{blue}{\underline{0.6184}} \\
        OSRT\cite{OSRT}        &23.82&0.6484&22.13&0.6301&20.58&0.6180 \\
        BPOSR\cite{BPOSR}       &23.75&0.6459&21.91&0.6282&20.23&0.6171 \\
        AdaIR\cite{AdaIR}       &\textcolor{blue}{\underline{23.85}}&\textcolor{blue}{\underline{0.6502}}
        &\textcolor{blue}{\underline{22.14}}&\textcolor{blue}{\underline{0.6306}}
        &\textcolor{blue}{\underline{20.59}}&\textcolor{red}{\textbf{0.6185}} \\
        D$^\text{2}$R$^\text{2}$OSR (Ours)
        &\textcolor{red}{\textbf{24.00}}&\textcolor{red}{\textbf{0.6528}}
        &\textcolor{red}{\textbf{22.16}}&\textcolor{red}{\textbf{0.6328}}
        &\textcolor{red}{\textbf{20.69}}&\textcolor{blue}{\underline{0.6184}} \\

        \bottomrule
    \end{tabular}
    \vspace{-4pt}
\end{table}


\section{Experiments}
\label{sec:expe}

\subsection{Implementation Details}

The proposed D$^\text{2}$R$^\text{2}$OSR presents a projection fusion representation framework for real-world ODI-SR.
We train our model on Flickr360 \cite{NTIRE2023}, and further evaluate it on Flickr360 \cite{NTIRE2023}, ODI-SR \cite{LAU-Net} and SUN360 \cite{SUN360}.
All HR ERP ODIs have a resolution of $1024\times2048$.
To simulate realistic degradations, We follow the degradation settings of Real-ESRGAN \cite{Real-ESRGAN}, while additionally incorporating fisheye and ERP-specific degradation distributions as indicated in Fig. \ref{fig:degradation}.
The network consists of $N=6$ EAMB-PAMB pairs, each with a depth of 6.
The channel dimensions are set to $c_e=180$ and $c_p=60$, respectively.
For DSMs, we set $k=5$, $c_{ed}=128$, and $c_{pd}=60$.
During training, HR ODIs are randomly cropped into 256 ${\times}$ 256 patches with random flip augmented. 
We train the model for 500K iterations using the Adam optimizer ($lr=1\times10^{-4}$, $\beta_1=0.9$, $\beta_2=0.99$) with an $\ell_1$ loss. 
The batch size is set to 4. All experiments are implemented in PyTorch and conducted on NVIDIA A6000 GPUs.

\subsection{Comparison Results}
To validate the effectiveness of D$^\text{2}$R$^\text{2}$OSR for real-world ODI-SR, we compare it with representative methods from three related domains: ODI-SR, Real-SR, and general image restoration.
The compared approaches include OSRT \cite{OSRT}, BPOSR \cite{BPOSR}, Real-ESRGAN \cite{Real-ESRGAN}, SwinIR \cite{SwinIR}, HAT \cite{HAT}, PromptIR \cite{PromptIR}, and AdaIR \cite{AdaIR}.
Following common practice in ODI evaluation, we report PSNR and SSIM together with ERP-specific metrics WS-PSNR \cite{WS-PSNR} and WS-SSIM \cite{WS-SSIM}, all computed on the luminance (Y) channel.
Since the parts of compared approaches are not originally designed for real-world ODI-SR, we retrain them under a unified training paradigm using our simulated real-world ODI-SR dataset.
All methods share identical training settings, ensuring that performance differences arise solely from architectural design differences.

\begin{figure*}[!t]
\centering
\includegraphics[width=0.95\columnwidth]{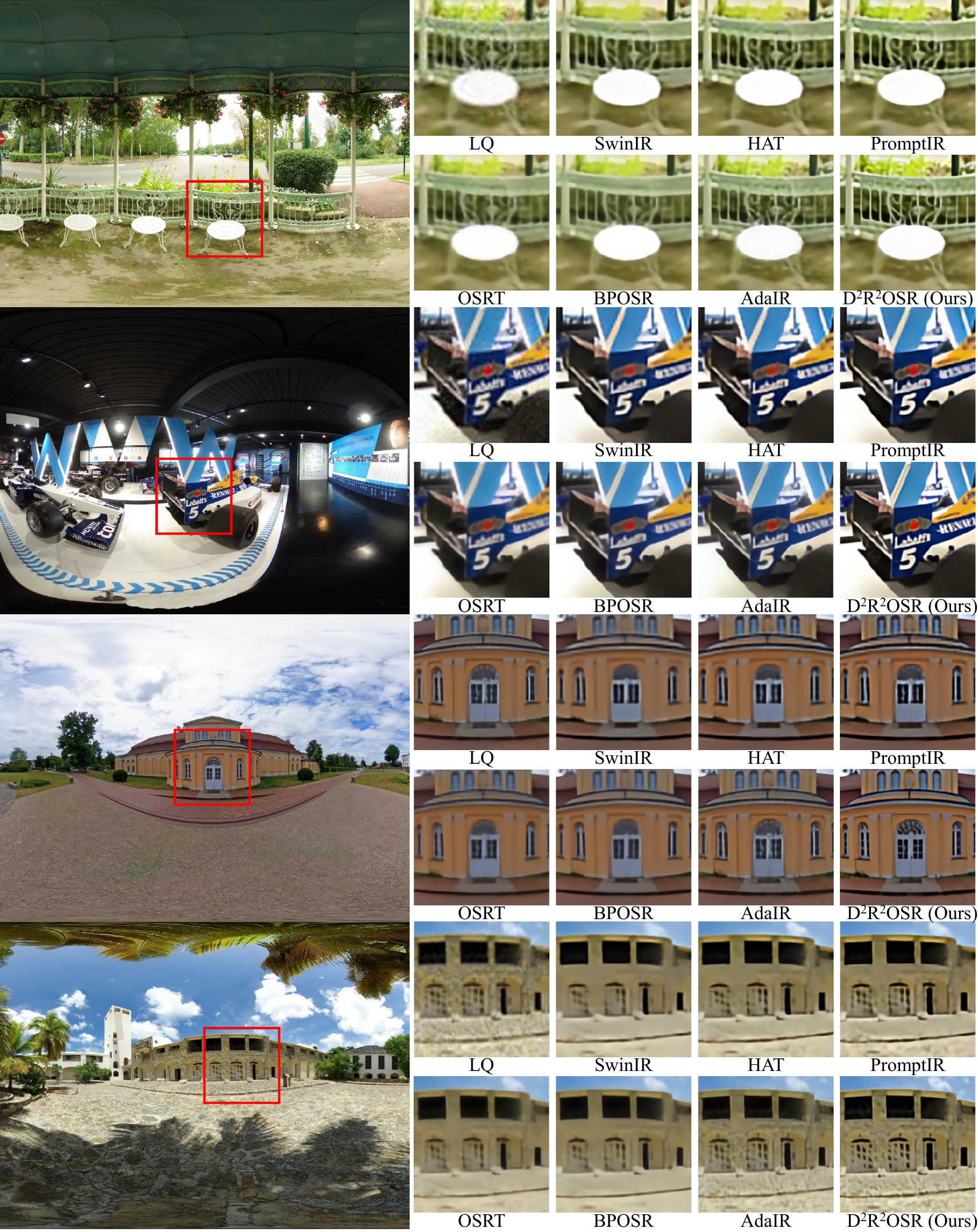}
\caption{Qualitative $\times$4 real-world ODI-SR comparison on ERP ODIs.}
\label{fig:SR}
\vspace{-24pt}
\end{figure*}

\subsubsection{Quantitative and Qualitative Comparison}
As shown in Tab. \ref{tab:SR}, D$^\text{2}$R$^\text{2}$OSR achieves the best quantitative performance across almost all benchmarks and scaling factors, validating the effectiveness of the proposed framework for real-world ODI-SR. 
In particular, D$^\text{2}$R$^\text{2}$OSR consistently outperforms the second-best approach AdaIR in WS-PSNR, with the largest gain of 0.28,dB achieved on Flickr360 under the $\times8$ setting.
Even under the extremely challenging $\times16$ setting, the proposed method maintains clear advantages on all three datasets, demonstrating strong robustness under severe degradations and large upscaling factors.
More quantitative comparisons of perceptual quality metrics with other methods are provided in the Appendix.


Fig. \ref{fig:SR} presents qualitative comparisons with representative ODI-SR and image restoration methods.
Upon inspection of zoomed-in regions, only D$^\text{2}$R$^\text{2}$OSR effectively suppresses severe artifacts while faithfully recovering fine details. 
For example, the chair structures exhibit more accurate textures and color consistency, while the picture frame in the second scene is reconstructed without introducing noticeable artifacts or noise.
These observations corroborate the superior capability of D$^\text{2}$R$^\text{2}$OSR in restoring fine structures and preserving visual fidelity.

\begin{table}[!t]
    \caption{Computational efficiency comparison on the SUN360 dataset.}
    \label{tab:effiency}
    
    \centering
    \footnotesize
    \setlength{\tabcolsep}{2pt}
    \begin{tabular}{c|ccccccccc}
    \toprule

    \multirow{2}{*}{Metric} &
    {\scriptsize\multirow{2}{*}{360-SS}} &
    {\scriptsize Real-} &
    {\scriptsize\multirow{2}{*}{SwinIR}} &
    {\scriptsize\multirow{2}{*}{HAT}} &
    {\scriptsize\multirow{2}{*}{PromptIR}} &
    {\scriptsize\multirow{2}{*}{OSRT}} &
    {\scriptsize\multirow{2}{*}{BPOSR}} &
    {\scriptsize\multirow{2}{*}{AdaIR}} &
    {\scriptsize D$^\text{2}$R$^\text{2}$OSR} \\
    
    &&{\scriptsize ESRGAN}&&&&&&&{\scriptsize (Ours)}\\
    
    \midrule
    
    Params. (M)$\downarrow$ & \textcolor{red}{\textbf{0.19}} &
    16.70 & 11.67 & 20.33 & 48.85 & 6.02 & \textcolor{blue}{\underline{2.07}} & 28.90 & 3.50 \\
    
    FLOPs (T)$\downarrow$ & 0.73 & 2.35 & 1.70 & 2.84 & 0.67 & 0.79 &
    \textcolor{red}{\textbf{0.32}} & 0.96 & \textcolor{blue}{\underline{0.55}} \\
    
    Runtime (s)$\downarrow$ & \textcolor{red}{\textbf{0.21}} & 0.69 &
    0.75 & 0.51 & 0.47 & 1.33 & 0.42 & 0.38 & \textcolor{blue}{\underline{0.34}} \\

    \bottomrule
    \end{tabular}
\end{table}

\begin{table}[!t]
\caption{Ablation studies for the parallel branch design and Perspective Projection Representation (PPR) on the real-world Flickr360 dataset.}
\label{tab:PPR}

\centering
\footnotesize
\renewcommand{\arraystretch}{1.0}
\setlength{\tabcolsep}{3.5pt}

\begin{tabular}{c|cc|ccc|cc}
\toprule

\multirow{2}{*}[-0.3em]{Method}&\multicolumn{2}{c|}{Branch}&\multicolumn{3}{c|}{Representation}&WS-PSNR$\uparrow$&PSNR$\uparrow$\\

\cmidrule(lr){2-6}

&Single&Dual&ERP&Pers&PPR&WS-SSIM$\uparrow$&SSIM$\uparrow$\\

\midrule

Baseline&\checkmark&&\checkmark&&&25.20 / 0.6858&24.63 / 0.6644\\
Perspective (Pers)&\checkmark&&&\checkmark&&25.09 / 0.6820&24.51 / 0.6602\\
Dual-Baseline&&\checkmark&\checkmark&&&25.35 / 0.6865&24.77 / 0.6650\\
Baseline + Pers&&\checkmark&\checkmark&\checkmark&&25.47 / 0.6882&24.90 / 0.6669\\
Baseline + PPR&&\checkmark&\checkmark&&\checkmark&25.50 / 0.6892&24.94 / 0.6682\\
\bottomrule
\end{tabular}
\end{table}

\subsection{Computational Efficiency}

Table~\ref{tab:effiency} compares the computational efficiency of different methods.
Runtime is measured on a single NVIDIA A6000 GPU with a batch size of 1 and an input resolution of $256\times512$ under the $\times4$ setting.
The proposed D$^\text{2}$R$^\text{2}$OSR achieves competitive efficiency with only 3.50M parameters, 0.55T FLOPs,, and an inference time of 0.34\,s per image. 
Compared with most Transformer-based methods, D$^\text{2}$R$^\text{2}$OSR provides a favorable balance between model complexity and inference speed.
In particular, it is approximately $4\times$ faster than OSRT (1.33, s/image) while achieving superior restoration performance.

\subsection{Ablation Studies}
In this section, we perform ablation studies to validate the effectiveness of each developed component.
Unless otherwise specified, all experiments are performed under the full D$^\text{2}$R$^\text{2}$OSR configuration for $\times4$ real-world ODI-SR.

\textbf{Effects of different dual-branch architectures and projection representations.}
Tab. \ref{tab:PPR} summarizes the impact of different branch designs and projection representations.
Directly projecting ERP ODIs into the perspective domain (second row) does not improve performance.
Simply introducing an additional branch by duplicating the network with reduced feature dimensions (third row) yields a 0.15 dB WS-PSNR improvement.
Compared with feature expansion within the same ERP distribution, incorporating a dedicated perspective branch (fourth row) provides complementary viewpoint information and further improves WS-PSNR by 0.27 dB.
Replacing conventional perspective projection (OpenCV Remap) with the proposed PPR (fifth row) further increases WS-PSNR to 25.50 dB.
These results validate the effectiveness of the proposed dual-branch design and the PPR representation for real-world ODI-SR.

\begin{table}[!t]
\caption{Round-trip reconstruction quality across different latitude regions.}
\label{tab:roundtrip}

\centering
\footnotesize
\renewcommand{\arraystretch}{1.0}
\setlength{\tabcolsep}{8pt}

\begin{tabular}{c|c|cc}
\toprule

\multirow{2}{*}{Viewpoint}&\multirow{2}{*}{Latitude} &\multicolumn{2}{c}{PSNR$\uparrow$ / SSIM$\uparrow$} \\

\cmidrule(lr){3-4}

&&OpenCV Remap&PPR (Ours) \\

\midrule

Equator&$0^\circ$&29.28 / 0.9056&34.89 / 0.9703 \\
Mid-lat 30&$30^\circ$&30.94 / 0.9233&36.27 / 0.9765 \\
Mid-lat 60&$60^\circ$&36.21 / 0.9404&40.07 / 0.9723 \\

\midrule

Overall&--&33.63 / 0.9295&38.17 / 0.9728 \\

\bottomrule
\end{tabular}

\end{table}

\begin{figure}[!t]
\centering
\includegraphics[width=\linewidth]{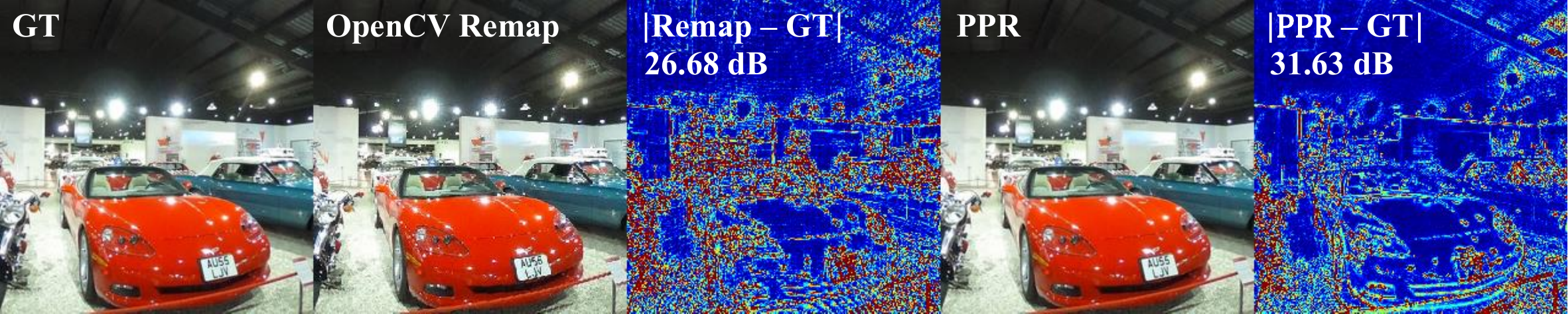}
\caption{Round-trip reconstruction visualization.}
\label{fig:ppr_compare}
\vspace{-5pt}
\end{figure}

\textbf{Geometric fidelity of PPR.}
To further validate the proposed PPR, we evaluate the round-trip ERP$\rightarrow$PPR$\rightarrow$ERP reconstruction quality on 1,200 training images across different latitude bands.
As reported in Tab.~\ref{tab:roundtrip}, PPR outperforms OpenCV Remap by an average of 4.54 dB PSNR, indicating substantially reduced projection-induced information loss across all latitudes.
Qualitative comparisons in Fig.~\ref{fig:ppr_compare} further show cleaner residual maps (brighter indicates larger reconstruction error) and better preservation of fine structures, such as the plate text ``AUSS''.
These results demonstrate the superior geometric fidelity of PPR and explain its consistent gains achieved by the dual-branch framework.

\begin{table}[!t]
\caption{Ablation studies of DSM and PFAM on the real-world Flickr360 dataset.}
\label{tab:DSM}

\centering
\footnotesize
\renewcommand{\arraystretch}{1.0}
\setlength{\tabcolsep}{5pt}

\begin{tabular}{c|cc}
\toprule

\multirow{2}{*}{Method}&WS-PSNR$\uparrow$&PSNR$\uparrow$ \\

&WS-SSIM$\uparrow$&SSIM$\uparrow$\\

\midrule

Baseline + PPR (Baseline$_P$)&25.50 / 0.6892&24.94 / 0.6682\\
Baseline$_P$ + $\text{DSM}_{\text{ERP}}$&25.56 / 0.6906&25.02 / 0.6702\\
Baseline$_P$ + $\text{DSM}_{\text{PPR}}$&25.58 / 0.6909&25.03 / 0.6703\\
Baseline$_P$ + $\text{DSM}_{\text{ERP} \& \text{PPR}}$ (Baseline$_D$)&25.58 / 0.6910&25.06 / 0.6712\\
Baseline$_D$ + Attention-Level Injection (ALI)&25.57 / 0.6918&25.08 / 0.6716\\
Baseline$_D$ + ALI + PFAM&25.64 / 0.6934&25.13 / 0.6726\\

\bottomrule
\end{tabular}

\end{table}

\textbf{Impact of individual architecture modules.}
We further evaluate the effectiveness of the proposed Degradation-Specific Module (DSM) and Projection Fusion Attention Module (PFAM) on the PPR-based dual-branch framework. 
As displayed in Tab.~\ref{tab:DSM}, introducing $\text{DSM}_{ERP}$ and $\text{DSM}_{PPR}$ improves WS-PSNR by 0.08\,dB, demonstrating the benefit of projection-specific degradation modeling.
Further incorporating attention-level injection (ALI) improves WS-PSNR to 25.57\,dB by enabling degradation-aware modulation within Transformer attention.
Finally, PFAM brings an additional 0.07\,dB gain through adaptive cross-branch fusion in the spatial-channel domain.
These results demonstrate that projection-specific degradation modeling and adaptive cross-projection fusion complement each other, jointly enabling D$^\text{2}$R$^\text{2}$OSR to effectively handle the complex degradations encountered in real-world ODIs.

\section{Conclusion}
\label{sec:con}

In this paper, we propose D$^\text{2}$R$^\text{2}$OSR, a Degradation-Disentangled Representation network for Real-world Omnidirectional image Super-Resolution in practical scenarios.
First, we model a realistic two-stage Fisheye-to-ERP degradation process based on the ODI imaging pipeline.
Motivated by two observations: (1) geometric projection degradations are entangled with random real-world distortions, and (2) users typically focus on limited viewpoints within the full 360$^{\circ}$ space, we develop a dual-branch architecture with Perspective Projection Representation (PPR) to explicitly separate degradations across different distributions while better aligning with human visual perception.
Moreover, we integrate Degradation-Specific Modules (DSMs) into both branches to adaptively extract degradation priors.
The enhanced features are further aggregated through Projection Fusion Attention Modules (PFAMs) at each stage to facilitate information interaction across projection domains.
Extensive experiments on multiple datasets demonstrate that D$^\text{2}$R$^\text{2}$OSR achieves state-of-the-art real-world ODI-SR performance while maintaining practical efficiency with fast inference for real-world viewpoint enhancement applications.

\section*{Acknowledgements}
This work was supported by the National Natural Science Foundation of China (Grant No. 62461160310, 62521007, 62431011), and the Fundamental Research Funds for the Central Universities (E2ET1104).

\newpage

%
%
\bibliographystyle{splncs04}
\bibliography{main}

@String(ECCV  = {Eur. Conf. Comput. Vis.})

@String(ICLR  = {Int. Conf. Learn. Represent.})

@String(AAAI  = {AAAI})

@String(ICIP  = {IEEE Int. Conf. Image Process.})

@String(ECCV  = {ECCV})

@String(ICLR  = {ICLR})

@String(ICIP  = {ICIP})

@inproceedings{SRCNN,
  title={Learning a deep convolutional network for image super-resolution},
  author={Dong, Chao and Loy, Chen Change and He, Kaiming and Tang, Xiaoou},
  booktitle={Computer Vision--ECCV 2014: 13th European Conference, Zurich, Switzerland, September 6-12, 2014, Proceedings, Part IV 13},
  pages={184--199},
  year={2014},
  organization={Springer}
}

@inproceedings{SwinIR,
  title={{SwinIR: Image restoration using swin transformer}},
  author={Liang, Jingyun and Cao, Jiezhang and Sun, Guolei and Zhang, Kai and Van Gool, Luc and Timofte, Radu},
  booktitle={Proceedings of the IEEE/CVF international conference on computer vision},
  pages={1833--1844},
  year={2021}
}

@inproceedings{RCAN,
  title={Image super-resolution using very deep residual channel attention networks},
  author={Zhang, Yulun and Li, Kunpeng and Li, Kai and Wang, Lichen and Zhong, Bineng and Fu, Yun},
  booktitle={Proceedings of the European conference on computer vision (ECCV)},
  pages={286--301},
  year={2018}
}

@inproceedings{HAT,
  title={Activating more pixels in image super-resolution transformer},
  author={Chen, Xiangyu and Wang, Xintao and Zhou, Jiantao and Qiao, Yu and Dong, Chao},
  booktitle={Proceedings of the IEEE/CVF conference on computer vision and pattern recognition},
  pages={22367--22377},
  year={2023}
}

@inproceedings{EDSR,
  title={Enhanced deep residual networks for single image super-resolution},
  author={Lim, Bee and Son, Sanghyun and Kim, Heewon and Nah, Seungjun and Mu Lee, Kyoung},
  booktitle={Proceedings of the IEEE conference on computer vision and pattern recognition workshops},
  pages={136--144},
  year={2017}
}

@inproceedings{DASR,
  title={Efficient and degradation-adaptive network for real-world image super-resolution},
  author={Liang, Jie and Zeng, Hui and Zhang, Lei},
  booktitle={European Conference on Computer Vision},
  pages={574--591},
  year={2022},
  organization={Springer}
}

@inproceedings{SRGAN,
  title={Photo-realistic single image super-resolution using a generative adversarial network},
  author={Ledig, Christian and Theis, Lucas and Husz{\'a}r, Ferenc and Caballero, Jose and Cunningham, Andrew and Acosta, Alejandro and Aitken, Andrew and Tejani, Alykhan and Totz, Johannes and Wang, Zehan and others},
  booktitle={Proceedings of the IEEE conference on computer vision and pattern recognition},
  pages={4681--4690},
  year={2017}
}

@inproceedings{DASR_,
  title={Unsupervised degradation representation learning for blind super-resolution},
  author={Wang, Longguang and Wang, Yingqian and Dong, Xiaoyu and Xu, Qingyu and Yang, Jungang and An, Wei and Guo, Yulan},
  booktitle={Proceedings of the IEEE/CVF conference on computer vision and pattern recognition},
  pages={10581--10590},
  year={2021}
}

@inproceedings{Real-ESRGAN,
  title={{Real-ESRGAN: Training real-world blind super-resolution with pure synthetic data}},
  author={Wang, Xintao and Xie, Liangbin and Dong, Chao and Shan, Ying},
  booktitle={Proceedings of the IEEE/CVF international conference on computer vision},
  pages={1905--1914},
  year={2021}
}

@inproceedings{RealBasicVSR,
  title={Investigating tradeoffs in real-world video super-resolution},
  author={Chan, Kelvin CK and Zhou, Shangchen and Xu, Xiangyu and Loy, Chen Change},
  booktitle={Proceedings of the IEEE/CVF Conference on Computer Vision and Pattern Recognition},
  pages={5962--5971},
  year={2022}
}

@inproceedings{DASR__,
  title={Unsupervised real-world image super resolution via domain-distance aware training},
  author={Wei, Yunxuan and Gu, Shuhang and Li, Yawei and Timofte, Radu and Jin, Longcun and Song, Hengjie},
  booktitle={Proceedings of the IEEE/CVF conference on computer vision and pattern recognition},
  pages={13385--13394},
  year={2021}
}

@inproceedings{BSRGAN,
  title={Designing a practical degradation model for deep blind image super-resolution},
  author={Zhang, Kai and Liang, Jingyun and Van Gool, Luc and Timofte, Radu},
  booktitle={Proceedings of the IEEE/CVF International Conference on Computer Vision},
  pages={4791--4800},
  year={2021}
}

@inproceedings{100viewpoint,
  title={Streaming 360-degree videos using super-resolution},
  author={Dasari, Mallesham and Bhattacharya, Arani and Vargas, Santiago and Sahu, Pranjal and Balasubramanian, Aruna and Das, Samir R},
  booktitle={IEEE INFOCOM 2020-IEEE Conference on Computer Communications},
  pages={1977--1986},
  year={2020},
  organization={IEEE}
}

@inproceedings{360-SS,
  title={Super-resolution of omnidirectional images using adversarial learning},
  author={Ozcinar, Cagri and Rana, Aakanksha and Smolic, Aljosa},
  booktitle={2019 IEEE 21st International Workshop on Multimedia Signal Processing (MMSP)},
  pages={1--6},
  year={2019},
  organization={IEEE}
}

@inproceedings{LAU-Net,
  title={{LAU-Net: Latitude adaptive upscaling network for omnidirectional image super-resolution}},
  author={Deng, Xin and Wang, Hao and Xu, Mai and Guo, Yichen and Song, Yuhang and Yang, Li},
  booktitle={Proceedings of the IEEE/CVF Conference on Computer Vision and Pattern Recognition},
  pages={9189--9198},
  year={2021}
}

@article{LAU-Net+,
  title={Omnidirectional image super-resolution via latitude adaptive network},
  author={Deng, Xin and Wang, Hao and Xu, Mai and Li, Li and Wang, Zulin},
  journal={IEEE Transactions on Multimedia},
  volume={25},
  pages={4108--4120},
  year={2022},
  publisher={IEEE}
}

@inproceedings{ICIP,
  title={Perception-Oriented Omnidirectional Image Super-Resolution Based on Transformer Network},
  author={An, Hongyu and Zhang, Xinfeng},
  booktitle={2023 IEEE International Conference on Image Processing (ICIP)},
  pages={3583--3587},
  year={2023},
  organization={IEEE}
}

@inproceedings{OSRT,
  title={{OSRT: Omnidirectional image super-resolution with distortion-aware transformer}},
  author={Yu, Fanghua and Wang, Xintao and Cao, Mingdeng and Li, Gen and Shan, Ying and Dong, Chao},
  booktitle={Proceedings of the IEEE/CVF Conference on Computer Vision and Pattern Recognition},
  pages={13283--13292},
  year={2023}
}

@inproceedings{FATO,
  title={{FATO: Frequency Attention Transformer for Omnidirectional Image Super-Resolution}},
  author={An, Hongyu and Zhang, Xinfeng and Zhao, Shijie and Zhang, Li},
  booktitle={Proceedings of the 6th ACM International Conference on Multimedia in Asia},
  pages={1--7},
  year={2024}
}

@inproceedings{ODA-SRN,
  title={Dual enhancement in ODI super-resolution: adapting convolution and upsampling to projection distortion},
  author={Ji, Xiang and Xu, Changqiao and Zhong, Lujie and Yang, Shujie and Xiao, Han and Muntean, Gabriel-Miro},
  booktitle={Proceedings of the Thirty-Third International Joint Conference on Artificial Intelligence},
  pages={902--910},
  year={2024}
}

@inproceedings{BPOSR,
  title={Omnidirectional image super-resolution via bi-projection fusion},
  author={Wang, Jiangang and Cui, Yuning and Li, Yawen and Ren, Wenqi and Cao, Xiaochun},
  booktitle={Proceedings of the AAAI Conference on Artificial Intelligence},
  pages={5454--5462},
  year={2024}
}

@inproceedings{NTIRE2023,
  title={{NTIRE 2023 challenge on 360$^{\circ}$  omnidirectional image and video super-resolution: Datasets, methods and results}},
  author={Cao, Mingdeng and Mou, Chong and Yu, Fanghua and Wang, Xintao and Zheng, Yinqiang and Zhang, Jian and Dong, Chao and Li, Gen and Shan, Ying and Timofte, Radu and others},
  booktitle={Proceedings of the IEEE/CVF conference on computer vision and pattern recognition},
  pages={1731--1745},
  year={2023}
}

@article{GDGT-OSR,
  title={Geometric distortion guided transformer for omnidirectional image super-resolution},
  author={Yang, Cuixin and Dong, Rongkang and Xiao, Jun and Zhang, Cong and Lam, Kin-Man and Zhou, Fei and Qiu, Guoping},
  journal={IEEE Transactions on Circuits and Systems for Video Technology},
  year={2025},
  publisher={IEEE}
}

@inproceedings{IPG,
  title={{Image processing GNN: Breaking rigidity in super-resolution}},
  author={Tian, Yuchuan and Chen, Hanting and Xu, Chao and Wang, Yunhe},
  booktitle={Proceedings of the IEEE/CVF conference on computer vision and pattern recognition},
  pages={24108--24117},
  year={2024}
}

@inproceedings{OmniSSR,
  title={{OmniSSR: Zero-shot omnidirectional image super-resolution using stable diffusion model}},
  author={Li, Runyi and Sheng, Xuhan and Li, Weiqi and Zhang, Jian},
  booktitle={European Conference on Computer Vision},
  pages={198--216},
  year={2024},
  organization={Springer}
}

@inproceedings{RealSR,
  title={Toward real-world single image super-resolution: A new benchmark and a new model},
  author={Cai, Jianrui and Zeng, Hui and Yong, Hongwei and Cao, Zisheng and Zhang, Lei},
  booktitle={Proceedings of the IEEE/CVF international conference on computer vision},
  pages={3086--3095},
  year={2019}
}

@article{AdaSR,
  title={Towards Real-World Super Resolution with Adaptive Self-Similarity Mining},
  author={Fan, Zejia and Yang, Wenhan and Guo, Zongming and Liu, Jiaying},
  journal={IEEE Transactions on Image Processing},
  year={2024},
  publisher={IEEE}
}

@article{ADL,
  title={Toward real-world super-resolution via adaptive downsampling models},
  author={Son, Sanghyun and Kim, Jaeha and Lai, Wei-Sheng and Yang, Ming-Hsuan and Lee, Kyoung Mu},
  journal={IEEE transactions on pattern analysis and machine intelligence},
  volume={44},
  number={11},
  pages={8657--8670},
  year={2021},
  publisher={IEEE}
}

@inproceedings{LWay,
  title={{Low-Res Leads the Way: Improving generalization for super-resolution by self-supervised learning}},
  author={Chen, Haoyu and Li, Wenbo and Gu, Jinjin and Ren, Jingjing and Sun, Haoze and Zou, Xueyi and Zhang, Zhensong and Yan, Youliang and Zhu, Lei},
  booktitle={Proceedings of the IEEE/CVF Conference on Computer Vision and Pattern Recognition},
  pages={25857--25867},
  year={2024}
}

@inproceedings{DKP,
  title={A dynamic kernel prior model for unsupervised blind image super-resolution},
  author={Yang, Zhixiong and Xia, Jingyuan and Li, Shengxi and Huang, Xinghua and Zhang, Shuanghui and Liu, Zhen and Fu, Yaowen and Liu, Yongxiang},
  booktitle={Proceedings of the IEEE/CVF conference on computer vision and pattern recognition},
  pages={26046--26056},
  year={2024}
}

@inproceedings{DiffBIR,
  title={{DiffBIR: Toward blind image restoration with generative diffusion prior}},
  author={Lin, Xinqi and He, Jingwen and Chen, Ziyan and Lyu, Zhaoyang and Dai, Bo and Yu, Fanghua and Qiao, Yu and Ouyang, Wanli and Dong, Chao},
  booktitle={European conference on computer vision},
  pages={430--448},
  year={2024},
  organization={Springer}
}

@inproceedings{SeeSR,
  title={{SeeSR: Towards semantics-aware real-world image super-resolution}},
  author={Wu, Rongyuan and Yang, Tao and Sun, Lingchen and Zhang, Zhengqiang and Li, Shuai and Zhang, Lei},
  booktitle={Proceedings of the IEEE/CVF conference on computer vision and pattern recognition},
  pages={25456--25467},
  year={2024}
}

@inproceedings{SUPIR,
  title={Scaling up to excellence: Practicing model scaling for photo-realistic image restoration in the wild},
  author={Yu, Fanghua and Gu, Jinjin and Li, Zheyuan and Hu, Jinfan and Kong, Xiangtao and Wang, Xintao and He, Jingwen and Qiao, Yu and Dong, Chao},
  booktitle={Proceedings of the IEEE/CVF conference on computer vision and pattern recognition},
  pages={25669--25680},
  year={2024}
}

@inproceedings{FAOR,
  title={Fast omni-directional image super-resolution: Adapting the implicit image function with pixel and semantic-wise spherical geometric priors},
  author={Shen, Xuelin and Wang, Yitong and Zheng, Silin and Xiao, Kang and Yang, Wenhan and Wang, Xu},
  booktitle={Proceedings of the AAAI Conference on Artificial Intelligence},
  pages={6833--6841},
  year={2025}
}

@inproceedings{LTEW,
  title={Learning local implicit fourier representation for image warping},
  author={Lee, Jaewon and Choi, Kwang Pyo and Jin, Kyong Hwan},
  booktitle={European Conference on Computer Vision},
  pages={182--200},
  year={2022},
  organization={Springer}
}

@inproceedings{ResVR,
  title={{ResVR: Joint rescaling and viewport rendering of omnidirectional images}},
  author={Li, Weiqi and Zhao, Shijie and Chen, Bin and Cheng, Xinhua and Li, Junlin and Zhang, Li and Zhang, Jian},
  booktitle={Proceedings of the 32nd ACM International Conference on Multimedia},
  pages={78--87},
  year={2024}
}

@article{WS-PSNR,
  title={Weighted-to-spherically-uniform quality evaluation for omnidirectional video},
  author={Sun, Yule and Lu, Ang and Yu, Lu},
  journal={IEEE signal processing letters},
  volume={24},
  number={9},
  pages={1408--1412},
  year={2017},
  publisher={IEEE}
}

@inproceedings{SUN360,
  title={Recognizing scene viewpoint using panoramic place representation},
  author={Xiao, Jianxiong and Ehinger, Krista A and Oliva, Aude and Torralba, Antonio},
  booktitle={2012 IEEE conference on computer vision and pattern recognition},
  pages={2695--2702},
  year={2012},
  organization={IEEE}
}

@inproceedings{LIIF,
  title={Learning continuous image representation with local implicit image function},
  author={Chen, Yinbo and Liu, Sifei and Wang, Xiaolong},
  booktitle={Proceedings of the IEEE/CVF conference on computer vision and pattern recognition},
  pages={8628--8638},
  year={2021}
}

@inproceedings{SRWarp,
  title={{SRWarp: Generalized image super-resolution under arbitrary transformation}},
  author={Son, Sanghyun and Lee, Kyoung Mu},
  booktitle={Proceedings of the IEEE/CVF conference on computer vision and pattern recognition},
  pages={7782--7791},
  year={2021}
}

@article{PromptIR,
  title={{PromptIR: Prompting for all-in-one image restoration}},
  author={Potlapalli, Vaishnav and Zamir, Syed Waqas and Khan, Salman H and Shahbaz Khan, Fahad},
  journal={Advances in Neural Information Processing Systems},
  volume={36},
  pages={71275--71293},
  year={2023}
}

@inproceedings{AdaIR,
  title={{AdaIR: Adaptive all-in-one image restoration via frequency mining and modulation}},
  author={Cui, Yuning and Zamir, Syed Waqas and Khan, Salman and Knoll, Alois and Shah, Mubarak and Khan, Fahad Shahbaz},
  booktitle={13th International Conference on Learning Representations, ICLR 2025},
  pages={57335--57356},
  year={2025},
  organization={International Conference on Learning Representations, ICLR}
}

@inproceedings{WS-SSIM,
  title={Weighted-to-spherically-uniform SSIM objective quality evaluation for panoramic video},
  author={Zhou, Yufeng and Yu, Mei and Ma, Hualin and Shao, Hua and Jiang, Gangyi},
  booktitle={2018 14th IEEE International Conference on Signal Processing (ICSP)},
  pages={54--57},
  year={2018},
  organization={IEEE}
}

@inproceedings{STDAN,
  title={Spatio-temporal distortion aware omnidirectional video super-resolution},
  author={An, Hongyu and Zhang, Xinfeng and Zhao, Shijie and Zhang, Li and Xiong, Ruiqin},
  booktitle={Proceedings of the AAAI Conference on Artificial Intelligence},
  pages={2309--2317},
  year={2026}
}
\end{document}